# Large Language Models Substantially Compress Well-Being Dispersion but Largely Preserve Its Socioeconomic Structure

Nattavudh Powdthavee, *Nanyang Technological University, Singapore*

***Corresponding author:** Nattavudh Powdthavee, School of Social Sciences, Nanyang Technological University, Singapore, 639818. Email: nick.powdthavee@ntu.edu.sg.

## Abstract

Research using large language models (LLMs) to generate synthetic populations shows that model outputs compress human variation. We ask whether this compression also distorts where dispersion is concentrated across social groups. Using 93,901 respondents from 66 countries and territories in Wave 7 of the World Values Survey, we compare observed life satisfaction with predictions from six LLMs. All six substantially understate overall dispersion. Yet after normalizing for these scale differences, they largely reproduce the human income gradient in well-being inequality: lower-income groups remain relatively more heterogeneous than higher-income groups. The pattern is robust to country fixed effects and alternative weighting schemes, and extends directionally to employment, education, and perceived control. In a separately preregistered mechanism experiment using Claude Sonnet 5, counterfactually increasing income strongly raises predicted mean life satisfaction but produces essentially no linear change in predicted dispersion. Removing income from the respondent profile leaves the negative dispersion gradient almost unchanged; the same pattern appears with Gemini and DeepSeek, although somewhat attenuated. These findings show that LLMs can substantially compress well-being dispersion while preserving much of its socioeconomic structure, and that this structure does not depend on the explicit income cue alone.

It is now well established that large language models (LLMs) tend to produce less heterogeneous responses than the humans they are intended to represent (1–5). Yet we know much less about whether, despite this compression, LLMs preserve the way heterogeneity is distributed across different groups in society. This distinction is important because, in principle, the compression of human variation may be less consequential for comparisons across groups if the statistical relationships present in the original population are preserved (6, 7). An LLM could substantially underestimate overall variation while still correctly capturing differences in average outcomes and differences in heterogeneity across groups. The relevant question is therefore not only how much heterogeneity an LLM preserves, but whether it preserves where that heterogeneity occurs. The concern is that an LLM could, for example, reproduce differences in average outcomes while making some populations appear much more homogeneous than they really are. Such errors could lead to misleading conclusions about the prevalence of extreme outcomes, who is most vulnerable, and who is most likely to benefit from an intervention, quantities that are often central to the evaluation and targeting of social policies (8–10).

Existing evidence suggests that LLMs can reproduce some features of human data while distorting others. In previous work on life satisfaction, LLMs recover the direction of established socioeconomic gradients in average well-being but often misestimate their magnitude (11). Across a much broader set of social science datasets, LLM-generated populations exhibit compressed marginal distributions while producing bivariate associations and predictive relationships that are often stronger than those observed in human data (4). Nevertheless, while existing works provide useful insight into first-moment fidelity, i.e., whether LLMs reproduce systematic differences in average outcomes across groups, they are largely silent on second-moment fidelity, i.e., whether LLMs also preserve systematic differences in the dispersion of outcomes across groups. Agreement in average responses does not necessarily imply agreement in the underlying response distributions (12). The combination of compressed marginal distributions and strong associations between variables has been described as "typological reasoning," whereby diverse individuals are represented as simplified social types (4). Statistical realism also appears to provide little indication of whether LLMs will accurately reproduce other quantities of interest: its relationship with the accuracy of LLM-generated treatment effects has been found to be weak (13). What remains unclear is whether LLMs preserve systematic differences in within-group heterogeneity after accounting for their general tendency to compress human variation. Specifically, compression may be approximately proportional to human heterogeneity across groups, leaving its structure intact, or it may be greater in more heterogeneous groups, thereby flattening the gradient itself.

Income and subjective well-being provide an ideal setting to examine this question. While the positive relationship between income and average well-being is both theoretically and empirically well established (14–18), income is also commonly found to be related to the distribution of well-being within the population. Well-being inequality, defined here as the dispersion of reported well-being within a population, has been found to decline with economic growth across countries and over time (19), while within populations, lower-income individuals display substantially greater well-being inequality than higher-income individuals (20). Other studies show that the relationship between income and well-being varies systematically across the well-being distribution, with income more strongly associated with well-being at some parts of the distribution than others (18, 21). If LLMs fail to reproduce this income-related structure in well-being inequality, they could misrepresent how well-being is distributed within income groups and, to the extent that these distributional relationships reflect causal processes, obscure important heterogeneity in who benefits from income-based interventions.

It is not a priori obvious whether LLMs should preserve this structure. LLMs can reproduce systematic relationships between demographic and socioeconomic characteristics and human responses, including established correlates of life satisfaction (11, 22, 23). They may therefore encode information not only about differences in typical outcomes across groups but also about differences in their distributions. Yet representing such distributional information and reproducing it through generated responses are distinct capabilities. LLMs have been found to describe population response distributions more accurately than they can simulate them (24), and to perform poorly when asked to generate samples from known statistical distributions (25). Providing richer person-specific information can also recover substantial amounts of otherwise missing between-person variation (26). General under-dispersion may therefore coexist with substantial fidelity in how dispersion varies across groups.

However, there are also reasons to expect the socioeconomic structure of heterogeneity to be distorted. LLMs tend to produce less diverse responses within demographic groups than humans do (27), and they have greater difficulty representing individuals whose views depart from those typically associated with their demographic characteristics (3, 26). Fidelity also varies systematically across demographic groups (28, 29). If these tendencies are more consequential in groups where human well-being is more heterogeneous, variation within those groups will be disproportionately compressed. LLMs could then reproduce the income gradient in average well-being while flattening, or even eliminating, the income gradient in well-being inequality.

We test whether LLMs preserve the income gradient in well-being inequality, measured by the standard deviation of life satisfaction, rather than simply whether they reproduce average well-being gradients or the overall level of human variation. Using Wave 7 of the World Values Survey (N = 93,901 individuals from 66 countries and territories) and following a preregistered study plan (https://aspredicted.org/sf62bj.pdf), we provide six leading LLMs with the demographic, socioeconomic, and attitudinal characteristics of individual respondents and ask them to predict each respondent's life satisfaction. We compare the income gradients in average life satisfaction and well-being inequality produced by the LLMs with those observed in the human data. Our primary test asks whether the income gradient in well-being inequality survives after accounting for each model's overall tendency to compress human variation. We then ask what accounts for any structural fidelity we observe. In a separately preregistered follow-up experiment using Claude Sonnet 5 (https://aspredicted.org/5hg8ba.pdf), we remove household income from respondents' profiles and, separately, counterfactually vary income while holding all other respondent characteristics fixed. This allows us to distinguish the role of the explicit income cue from information contained in the broader respondent profile. We additionally replicate the income-removal analysis using Gemini 3.8 Flash and DeepSeek V4.1 Flash to examine whether the result extends beyond Claude.

## Results

### Income gradients in average well-being and well-being inequality

We first examined the relationship between household income position and the distribution of life satisfaction in the human data; see Fig. 1. Household income position is measured in the World Values Survey as respondents' self-reported position in their country's income distribution, from 1 (lowest) to 10 (highest), and is therefore already normalized within country. Panel A shows that average life satisfaction rises with household income position, in line with previous evidence on the positive relationship between income and subjective well-being (14–18). A one-group increase in household income position was associated with a 0.217-point increase in life satisfaction (95% CI, 0.176 to 0.263).

Panel B shows that well-being inequality, measured by the standard deviation of life satisfaction, generally declines with household income position, consistent with previous evidence on income-related differences in well-being inequality (19, 20). The linear slope across the ten income groups was −0.106 (95% CI, −0.126 to −0.083). Well-being inequality is highest at the bottom of the income distribution, declines across the lower income groups, remains comparatively flat through much of the middle and upper-middle range, and rises again at the highest income group. Overall, the pattern shows greater heterogeneity in well-being at lower income levels, despite the rebound in inequality at the top of the income distribution.

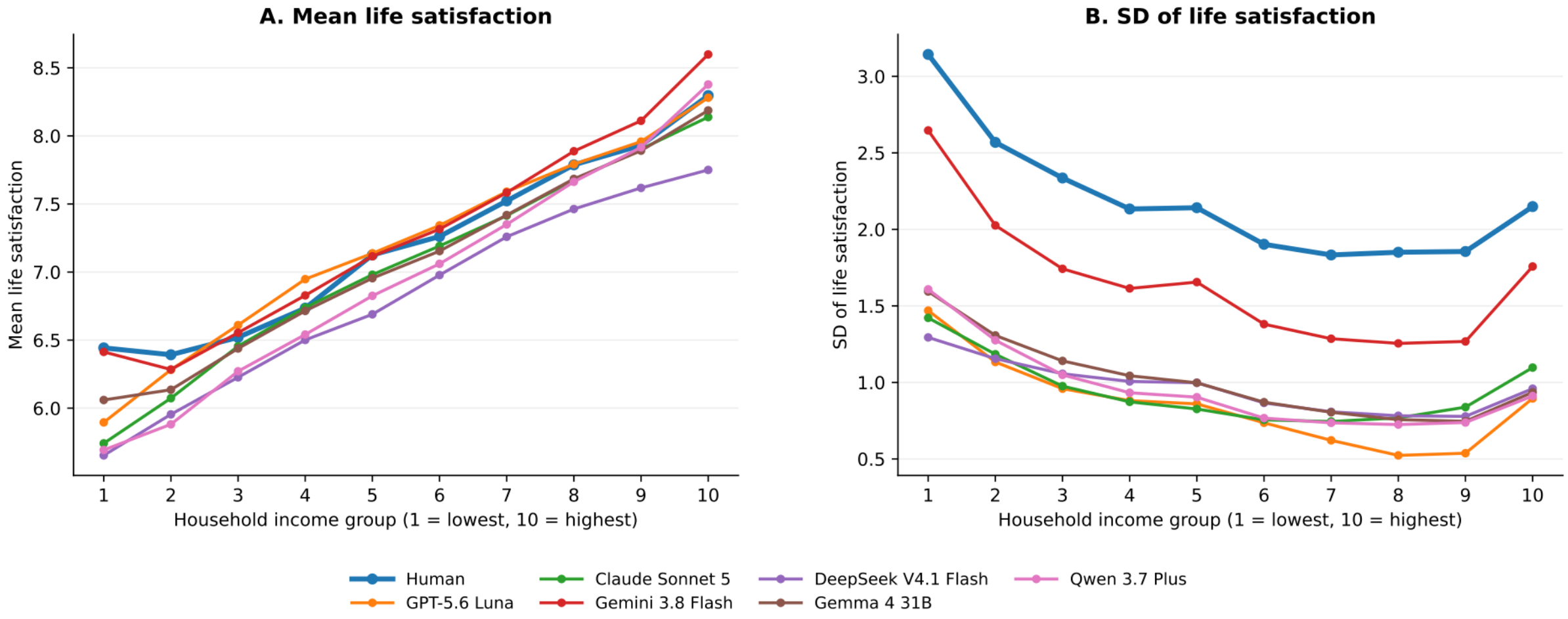

**Fig. 1. Income gradients in average well-being and well-being inequality in humans and large language models.** Panel A shows mean life satisfaction by household income group for human respondents and predictions from each of the six LLMs. Panel B shows the corresponding standard deviation of life satisfaction within each income group. Household income position is measured using World Values Survey item Q288, ranging from 1 (lowest income group) to 10 (highest income group). Human life satisfaction is the observed response to Q49; LLM life satisfaction is the respondent-level prediction generated from the corresponding WVS profile. All analyses use the same sample of 93,901 respondents from 66 countries and territories. Panel B reports raw, unnormalized standard deviations. Lines connect income-group estimates for visual presentation.

Turning to the LLM predictions, all six models reproduced the positive income gradient in average life satisfaction shown in Fig. 1A. The estimated slopes were 0.256 for GPT-5.6 Luna (95% CI, 0.237 to 0.275), 0.262 for Claude Sonnet 5 (95% CI, 0.241 to 0.286), 0.240 for Gemini 3.8 Flash (95% CI, 0.209 to 0.270), 0.249 for DeepSeek V4.1 Flash (95% CI, 0.229 to 0.268), 0.239 for Gemma 4 31B (95% CI, 0.213 to 0.266), and 0.284 for Qwen 3.7 Plus (95% CI, 0.259 to 0.310). All six models therefore reproduced the positive human income gradient in average well-being, although each estimated a somewhat steeper gradient than observed in the human data. This pattern is consistent with previous evidence that LLMs can recover systematic relationships between socioeconomic characteristics and human responses while misestimating their magnitude (11, 22, 23).

A more relevant question is whether LLMs preserve the structure of the income gradient in well-being inequality. Fig. 1B shows substantial compression in the absolute level of well-being inequality, consistent with previous evidence that LLM-generated responses tend to be less heterogeneous than human responses (1, 2, 4, 5). To summarize overall within-income inequality, we averaged the income-group-specific standard deviations, weighted by the human sample share in each income group, and applied the same weights to the human data and to every LLM. This provides a common measure of the overall scale of well-being inequality for each source while holding the income-group composition fixed. This weighted average was 2.168 in the human data, compared with 0.864 for GPT-5.6 Luna, 0.898 for Claude Sonnet 5, 1.642 for Gemini 3.8 Flash, 0.977 for DeepSeek V4.1 Flash, 1.021 for Gemma 4 31B, and 0.945 for Qwen 3.7 Plus. Relative to the human benchmark, this corresponds to compression of approximately 60% for GPT, 59% for Claude, 24% for Gemini, 55% for DeepSeek, 53% for Gemma, and 56% for Qwen.

Despite this compression, all six models show a statistically significant negative raw income gradient in well-being inequality. The estimated slopes were −0.075 for GPT-5.6 Luna (95% CI, −0.084 to −0.066), −0.041 for Claude Sonnet 5 (95% CI, −0.054 to −0.029), −0.103 for Gemini 3.8 Flash (95% CI, −0.120 to −0.084), −0.047 for DeepSeek V4.1 Flash (95% CI, −0.059 to −0.035), −0.076 for Gemma 4 31B (95% CI, −0.085 to −0.066), and −0.075 for Qwen 3.7 Plus (95% CI, −0.084 to −0.065). The raw profiles thus suggest that the models retain at least some of the human income-related structure in well-being inequality even while substantially compressing its absolute level.

**Normalized income gradients in well-being inequality**

However, Fig. 1B may be misleading. Because the LLMs differ substantially from humans in overall well-being inequality, the raw slopes partly reflect differences in scale rather than differences in the shape of the income gradient itself. For example, if human standard deviations across two income groups were 3 and 2, while an LLM produced 1.5 and 1, the LLM would have compressed well-being inequality by 50% at both income levels while preserving the relative difference between them exactly. Its raw gradient would nevertheless appear only half as large.

To separate overall compression from the structure of the income gradient, Fig. 2 normalized each income-group-specific standard deviation by the source's weighted average standard deviation across income groups. More specifically, for income group $j$ and source $m$,

$$D_{jm} = \frac{SD_{jm}}{\sum_{k=1}^{10} w_k SD_{km}},$$

where $SD_{jm}$ is the standard deviation of life satisfaction for income group $j$ from source $m$, and $w_k$ is the human sample share in income group $k$. The same human income-group weights are used for the human data and for every LLM. A normalized value above 1 indicates that well-being inequality in that income group is higher than the source's overall weighted average, whereas a value below 1 indicates lower-than-average inequality.

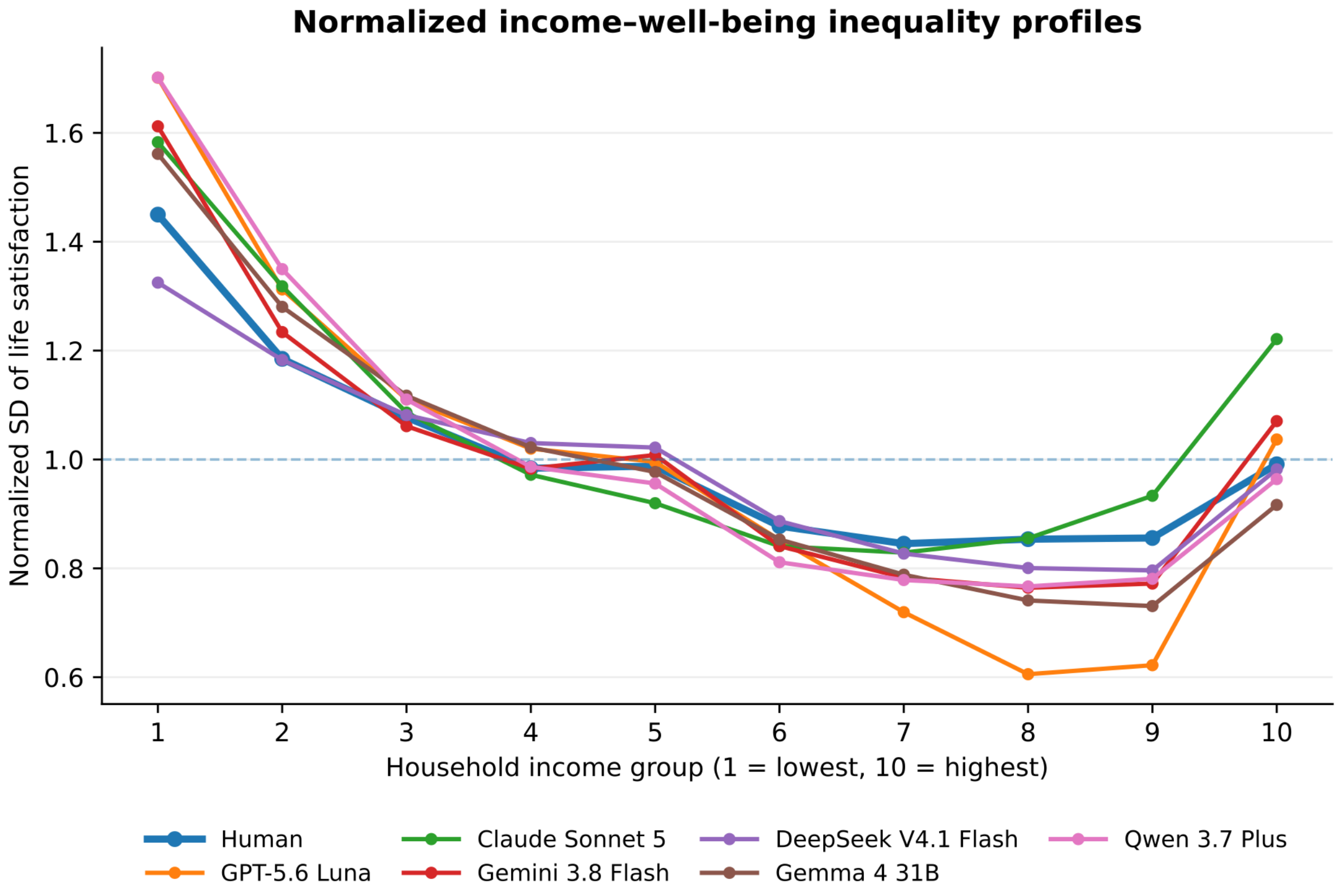

**Fig. 2. Normalized income profiles of well-being inequality in humans and large language models.** The figure shows the standard deviation of life satisfaction within each household income group after normalizing each source by its own weighted-average standard deviation across the ten income groups. The same human income-group shares are used as weights for humans and all six LLMs. A value of 1 therefore represents the source-specific weighted average; values above or below 1 indicate relatively greater or lower well-being inequality within that income group. This normalization removes differences in overall dispersion scale and allows comparison of the relative income profile of heterogeneity across sources. Household income position is measured using WVS item Q288.

We can see from Fig. 2 that much of the human income gradient in well-being inequality remains after these differences in scale are removed. All six LLMs place relatively greater well-being inequality at the bottom of the income distribution, show a broad decline as income rises, and reproduce the increase at the highest income group. Take Claude Sonnet 5, for example. At the bottom of the income distribution, its predicted well-being inequality is approximately 58% higher than Claude's own weighted average across the ten income groups. It falls steadily with income, reaching approximately 17% below its weighted average at income group 7, before rising again to approximately 22% above its weighted average at income group 10. The corresponding human values are about 45% above the human weighted average at income group 1, 15% below it at income group 7, and approximately equal to it at income group 10.

The steepness of the normalized gradient nevertheless differs across models. Relative to the human normalized slope, GPT-5.6 Luna produced a steeper gradient by −0.038 (95% CI, −0.047 to −0.030; $p < .001$), Gemini 3.8 Flash by −0.014 (95% CI, −0.021 to −0.006; $p < .001$), Gemma 4 31B by −0.026 (95% CI, −0.033 to −0.018; $p < .001$), and Qwen 3.7 Plus by −0.030 (95% CI, −0.038 to −0.023; $p < .001$). By contrast, Claude Sonnet 5 differed from the human slope by +0.003 (95% CI, −0.006 to 0.013; $p = .582$), and DeepSeek V4.1 Flash by +0.001 (95% CI, −0.007 to 0.009; $p = .834$).

Similarity in the linear gradient does not, however, imply that the full income profiles are identical. The preregistered global tests of equality between each LLM's normalized profile and the human profile rejected exact equality for all six models (Table S1). Hence, we do not interpret the LLMs as reproducing the human profile exactly. Rather, the normalized profiles show that, despite these statistically detectable differences, the models preserve much of the human profile's broad socioeconomic structure, including greater relative heterogeneity at lower income positions. This is our main finding.

In addition, the normalized slope differences changed little when countries were given equal weight or when WVS survey weights were applied (Table S2). We also conducted exploratory analyses of distributional tails and found that the models broadly recovered where very low and very high life satisfaction were concentrated across the income distribution, although they often misestimated the corresponding prevalence rates (Fig. S1).

**Robustness to country and demographic composition**

One objection to Fig. 2's findings is that the income gradient in well-being inequality may reflect country-specific differences rather than a relationship that holds within countries. We therefore estimate recentered influence function (RIF) regressions of life-satisfaction variance on household income position with country fixed effects. This allows us to test whether the gradient remains statistically significantly different from zero within countries.

**Table 1. Within-country income gradients in well-being inequality**

| Source | Income slope | SE | p-value |
|---|---|---|---|
| Human | −0.525 | 0.051 | <0.001 |
| GPT-5.6 Luna | −0.216 | 0.030 | <0.001 |
| Claude Sonnet 5 | −0.165 | 0.032 | <0.001 |
| Gemini 3.8 Flash | −0.424 | 0.046 | <0.001 |
| DeepSeek V4.1 Flash | −0.131 | 0.023 | <0.001 |
| Gemma 4 31B | −0.197 | 0.030 | <0.001 |
| Qwen 3.7 Plus | −0.196 | 0.036 | <0.001 |

**Note.** Entries are coefficients from recentered influence function (RIF) regressions of life-satisfaction variance on household income group (Q288), with country fixed effects. Negative coefficients indicate a negative association between household income position and the variance RIF after accounting for country fixed effects. Standard errors are clustered by country.

The income gradient remained negative for humans and for all six LLMs (Table 1). For humans, the coefficient was −0.525 (SE = 0.051; $p < .001$). The corresponding coefficients were −0.216 for GPT-5.6 Luna, −0.165 for Claude Sonnet 5, −0.424 for Gemini 3.8 Flash, −0.131 for DeepSeek V4.1 Flash, −0.197 for Gemma 4 31B, and −0.196 for Qwen 3.7 Plus (all $p < .001$). Thus, the negative income gradient in well-being inequality is not solely due to differences in country composition.

As a supplementary check, we also used RIF-Oaxaca decompositions to examine whether the most distinctive changes in life-satisfaction variance across adjacent income groups could be attributed to differences in age, sex, marital status, education, number of children, and country composition. For both the sharp decline between income groups 1 and 2 and the rebound between groups 9 and 10, most of the change remained in the residual component for humans and all six LLMs; full results for all adjacent-income comparisons are reported in Table S3.

**What generates the preserved income-dispersion structure?**

There are at least two potential mechanisms underlying the preserved dispersion gradient. One possibility is that the model responds directly to the respondent's stated income position, perhaps because its training data encode the documented relationship between income and the distribution of subjective well-being (18-21), thereby producing different levels of predicted dispersion across income levels. Another possibility is that the gradient arises from the broader configuration of respondent characteristics correlated with income, such as age, education, marital status, employment, occupation, and material hardship. In the original profiles, these sources of information appear together, so the main analysis cannot distinguish between them.

To distinguish between these mechanisms, we conducted a separately preregistered follow-up experiment using a stratified random sample of 2,000 respondents, with 200 from each of the ten income groups, and Claude Sonnet 5. Claude provided a particularly informative case because its normalized linear income-dispersion slope was almost identical to the human slope despite substantial compression in absolute dispersion.

First, we assigned each respondent all ten possible income positions in turn ($N = 2,000 \times 10 = 20,000$ profiles), changing only the income cue while leaving every other element of that respondent's profile unchanged. A negative gradient in this counterfactual sweep would indicate that changing income alone was sufficient to generate the pattern. Second, we removed household income from each respondent's profile while leaving all other characteristics unchanged. If the gradient persisted, this would indicate that the broader profile was sufficient to reproduce it. We later repeated the income-ablation analysis using Gemini 3.8 Flash and DeepSeek V4.1 Flash to assess whether the result extended beyond Claude.

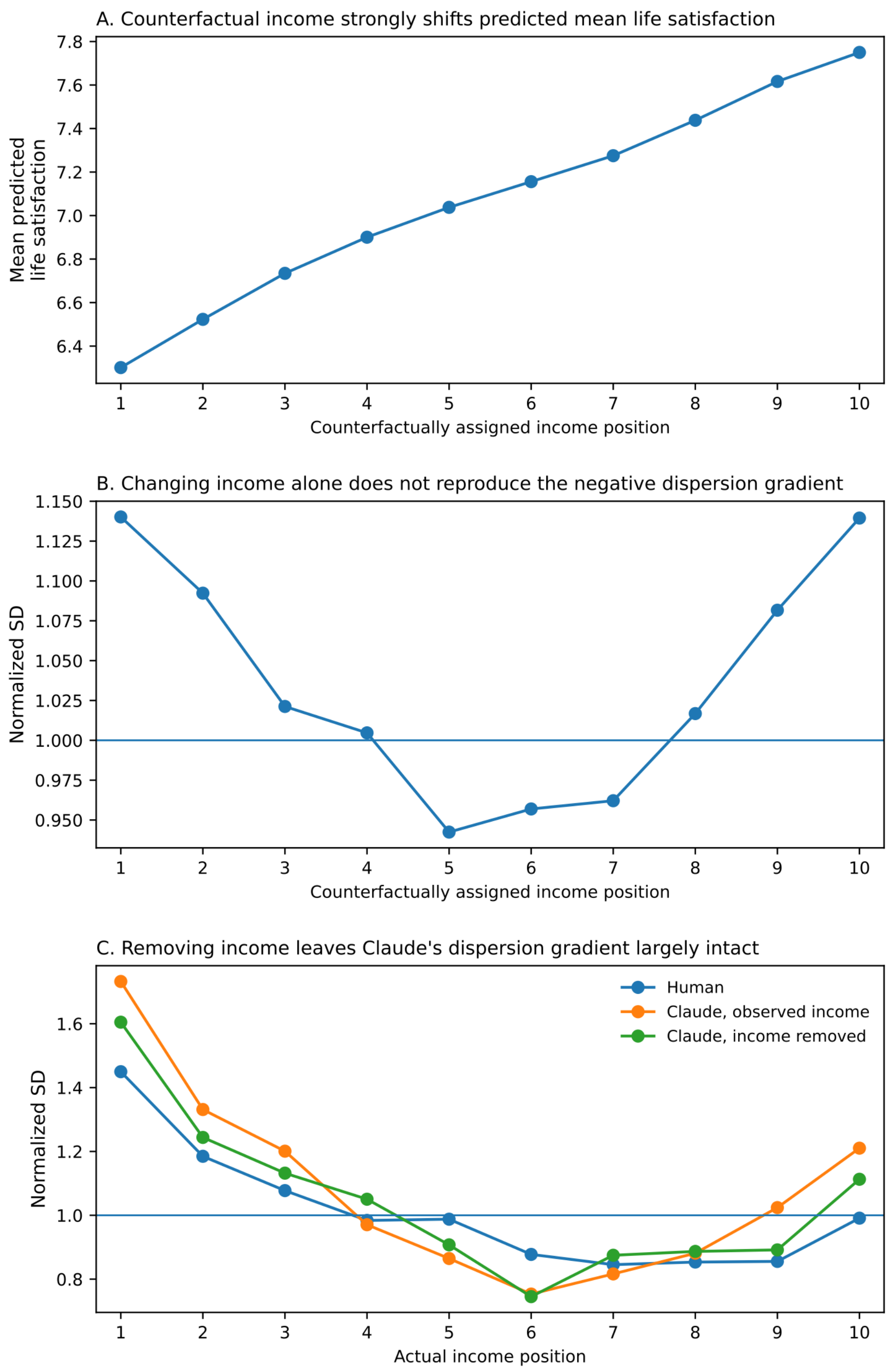

**Fig 3. Counterfactual and income-ablation tests of the income-related structure of predicted well-being heterogeneity.** The analyses use a stratified random sample of 2,000 WVS respondents, with 200 drawn from each income group. In Panels A and B, each respondent is assigned all ten possible income positions in turn while every other profile characteristic remains unchanged. Panel A shows mean predicted life satisfaction; Panel B shows normalized dispersion. Panel C compares the human dispersion profile with Claude's predictions when income is included versus removed. Normalized standard deviations use the same human income-group weights as in the main analysis.

Figure 3 reports the results of the mechanism experiment. Panel A shows that Claude responded strongly to the counterfactual income manipulation. Holding every other profile characteristic fixed within respondents, predicted life satisfaction increased by 0.154 points for each one-unit increase in assigned income position (95% CI, 0.148 to 0.161; $p < .001$). Thus, when asked to predict average life satisfaction, the model was clearly sensitive to the income information supplied in the profile.

The effect of assigned income on predicted dispersion was very different. In Panel B, the linear slope of normalized dispersion across the ten counterfactual income positions was −0.001 (95% CI, −0.006 to 0.004; $p = .628$), providing no evidence that changing income alone generates the negative dispersion gradient observed in the human data. The full profile was not flat, however. Dispersion followed a U-shaped pattern, with a positive quadratic term of 0.009 (95% CI, 0.008 to 0.011; $p < .001$); average dispersion at the two endpoints was approximately 21% higher than at the minimum (95% CI, 18% to 26%). The income cue therefore affects the distribution of Claude's predictions, but does not by itself reproduce the approximately declining income-dispersion relationship found in the observed data.

Panel C presents the complementary income-ablation test. With income included in the profile, Claude's normalized dispersion slope was −0.055 (95% CI, −0.080 to −0.026). After household income position was removed entirely, the slope remained −0.054 (95% CI, −0.076 to −0.030). The difference between the two slopes was −0.001 (95% CI, −0.013 to 0.010; $p = .878$). Thus, removing explicit income information leaves the broad income-related structure of Claude's predicted heterogeneity largely intact. We found a similar pattern when we repeated the income-ablation analysis using Gemini 3.8 Flash and DeepSeek V4.1 Flash (Fig. S2). In both models, the negative dispersion gradient remained after income was removed, though it became somewhat flatter. Based on these results, we conclude that the negative dispersion cannot be explained solely by the explicit income cue. Rather, much of the negative income gradient in well-being dispersion can be reproduced from information elsewhere in the respondent profile.

**Exploratory country-level gradient fidelity**

We next examine whether the structural fidelity observed in the pooled data extends across countries (Fig. 4). Specifically, we ask whether countries with steeper human income gradients in well-being inequality also tend to exhibit steeper gradients in the LLM predictions. Iraq is the only clear outlier in this analysis: in the WVS data, income and life satisfaction are nearly identical for many Iraqi respondents, producing an anomalously large positive human gradient. We thus exclude Iraq from Fig. 4 and the main country-level analysis, and report the full 66-country results, including Iraq, in Fig. S3 in the SI.

Across the remaining 65 countries and territories, country-specific human and LLM gradients were positively related for all six models. Rank correspondence was relatively stable across models, with Spearman correlations ranging from 0.54 to 0.80. Pearson correlations ranged from 0.51 to 0.79 after excluding Iraq, but were substantially lower when Iraq was retained because of its extreme human gradient (Fig. S3). The fitted slopes ranged from 0.71 to 1.04, with Gemini showing the closest correspondence to the human cross-country pattern in the Iraq-excluded specification. Overall, the models recover some of the cross-country ordering in the strength of the income-dispersion gradient, but the magnitude of linear correspondence is sensitive to influential country-level observations.

These results suggest that LLMs recover some of the cross-country ordering in the strength of the income-dispersion gradient, but correspondence in absolute gradient magnitudes is less robust. Hence, it appears that a model can reproduce the pooled income pattern reasonably well without closely reproducing the gradient in every country.

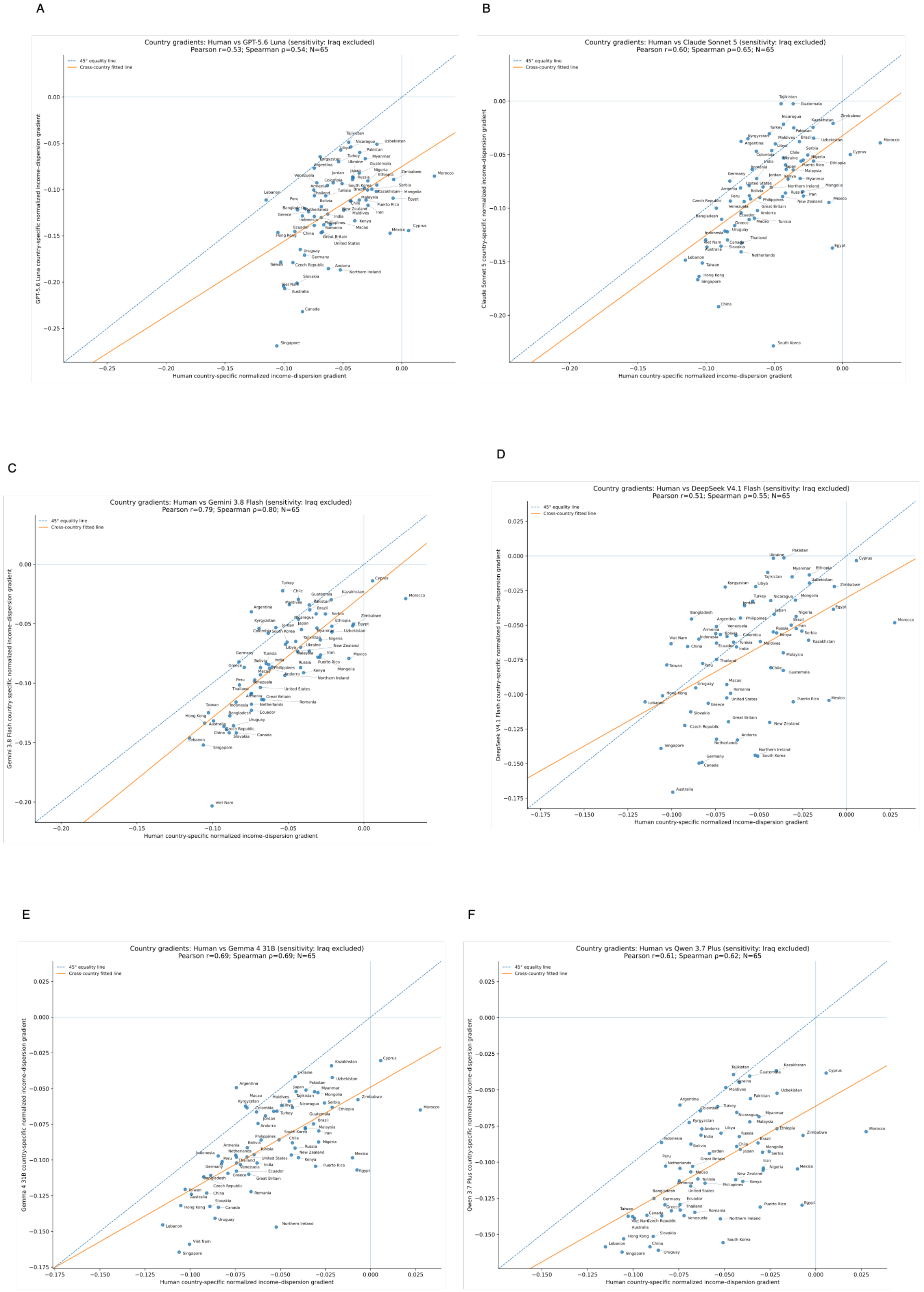

**Fig 4. Country-level fidelity in income gradients in well-being inequality.** Each panel compares the country-specific human income gradient in normalized life-satisfaction dispersion with the corresponding gradient produced by one LLM. Country-specific gradients are estimated from the relationship between household income

position and normalized within-income standard deviations of life satisfaction. Each point represents one country or territory. The dashed 45° line indicates equality between the human and LLM gradients, and the solid fitted line shows the cross-country linear relationship. Iraq is excluded from the main figure because it is a clear outlier in the human country-specific gradient; results including Iraq are reported in Fig. S3. The remaining sample contains 65 countries and territories.

## Generalization beyond income

One possible objection is that the preservation of structured well-being inequality may be specific to income rather than generalizing to other socioeconomic and psychosocial characteristics. To test this, we examine, as an exploratory analysis, whether the preservation of structured well-being inequality extends beyond income. We repeat the same scale-adjusted comparisons for employment status, education, marital status, and perceived freedom and control, with the full profiles reported in Fig. S4 in the SI.

The human data show greater well-being inequality among unemployed than employed respondents, among those with lower rather than tertiary education, and among those reporting low rather than high perceived control. The corresponding SD ratios are 1.25 for unemployment (95% CI, 1.17 to 1.31), 1.23 for lower education (95% CI, 1.17 to 1.30), and 1.45 for low perceived control (95% CI, 1.35 to 1.56). All six LLMs reproduce the direction of each difference. For unemployment, the ratios range from 1.12 for DeepSeek V4.1 Flash (95% CI, 1.08 to 1.16) to 1.38 for Qwen 3.7 Plus (95% CI, 1.33 to 1.43). For perceived control, they range from 1.22 for DeepSeek (95% CI, 1.15 to 1.30) to 1.84 for GPT-5.6 Luna (95% CI, 1.71 to 1.97). The marital-status difference is much smaller in the human data, with an SD ratio of 1.03 for non-partnered relative to partnered respondents (95% CI, 1.01 to 1.06).

The same patterns remain after accounting for country differences. In country-fixed-effect RIF regressions, unemployment is associated with greater life-satisfaction variance among humans (1.787; 95% CI, 1.350 to 2.225; $p < .001$), and the corresponding coefficient is positive for all six LLMs. Human life-satisfaction variance also declines with education (−0.308; 95% CI, −0.373 to −0.242; $p < .001$) and perceived control (−0.488; 95% CI, −0.603 to −0.373; $p < .001$), with all six models reproducing both directions. These results suggest that LLMs preserve much of the structure of well-being inequality across several socioeconomic and psychosocial characteristics, not only across income groups.

Finally, as a post-preregistration benchmark, we estimated out-of-sample OLS and Lasso predictions using observed WVS life satisfaction as the training outcome. Both models compressed overall dispersion but retained much of the normalized income-dispersion profile (Table S4). Their profile RMSDs from the human benchmark were 0.081 for OLS and 0.079 for Lasso, compared with 0.051 for DeepSeek, 0.075 for Gemini, 0.079 for Gemma, 0.100 for Claude, 0.107 for Qwen, and 0.147 for GPT. These results suggest that compression is not uniquely an LLM phenomenon: point predictors can generally understate dispersion. What differs across predictive systems is how much of the structure of human heterogeneity they preserve. The supervised models are not directly comparable to the LLMs, however, because OLS and Lasso were trained on observed life-satisfaction outcomes from the WVS, whereas the LLMs were not given life-satisfaction outcomes from the study sample and nevertheless recovered much of the income-related structure in human heterogeneity. The fact that the same combination of compression and structural preservation also appears in these supervised point predictors is consistent with the possibility that second-moment structure can emerge from individual-level prediction without being represented as an explicit group-level dispersion relationship.

## Discussion

Large language models tend to produce responses that are less heterogeneous than those observed in human populations (1–5). This has raised serious doubts about using them to simulate people or construct synthetic populations, particularly when the quantities of interest depend on within-group variation rather than average outcomes alone (3–5, 24–27). Our results show that this concern needs qualification. All six models substantially compress the overall distribution of life satisfaction among 93,901 individuals from 66 countries and territories in Wave 7 of the World Values Survey, yet they still reproduce much of the socioeconomic structure of well-being inequality. They place relatively higher and lower heterogeneity in broadly the same parts of the income distribution as humans, and similar patterns emerge across several other socioeconomic and psychosocial characteristics. This is the study's core finding: an LLM can substantially compress variation in human experiences while still providing insight into where that variation is concentrated across different socioeconomic and psychosocial characteristics.

Follow-up analyses with Claude as the primary model clarify the source of this preserved structure. Changing the income cue substantially shifted Claude's predicted mean life satisfaction but did not reproduce the negative linear dispersion gradient observed in the human data. By contrast, removing income from the respondent profile left Claude's negative dispersion gradient largely unchanged. Gemini and DeepSeek showed the same pattern, although their gradients became somewhat flatter. This suggests that the first- and second-moment relationships arise differently: predicted mean life satisfaction responds strongly to the explicit income cue, whereas the preserved income gradient in dispersion does not depend solely on that cue. Instead, the second-moment structure may emerge from how the model maps heterogeneous combinations of respondent characteristics into point predictions, rather than from a direct income-conditioned representation of how dispersion itself should vary across income groups. Our mechanism tests thus provide new insight into the synthetic-LLM literature by showing that distributional structure can emerge from individual-level prediction, even when the model is not explicitly prompted to reproduce group-level dispersion. If this mechanism applies more broadly, a similar second-moment structure may also emerge for other outcomes whenever the respondent profile contains characteristics that systematically organize variation in those outcomes.

Our findings have important implications for how we evaluate synthetic LLM data in the social sciences. Because raw LLM variation is distorted by systematic compression, absolute differences in heterogeneity across groups can be difficult to interpret directly. Normalizing each model by its own overall dispersion helps separate this scale distortion from the relative structure of heterogeneity. For exploratory policy targeting, this may provide a diagnostic of where LLM-generated heterogeneity is concentrated and where human validation may be especially important. This is relevant to policy design, where identifying meaningful subgroup variation is central to deciding who should receive an intervention (8–10). Normalization does not, however, recover the true magnitude of human heterogeneity. LLM-generated data may therefore help indicate where heterogeneous outcomes are concentrated while still providing poor estimates of the size of those differences, the prevalence of vulnerability, or the expected gains from intervention. This conclusion aligns with recent evidence that LLMs can reproduce the relative ordering of experimental treatment effects while misestimating their magnitude (30).

The paper's results also suggest that first-moment and second-moment fidelity in LLM research should be evaluated separately. Much of the literature on LLM-generated synthetic data has focused on whether models reproduce individual responses, average outcomes, or average differences across groups and populations (22, 23, 31, 32). Our results indicate that this leaves an important part of statistical fidelity unexamined. For example, second-moment fidelity in our setting concerns whether the dispersion of one generated prediction per respondent profile, under the generation settings used here, reproduces the corresponding dispersion observed in human data. Future work should therefore examine not only means and associations, but also how well LLM-generated data reproduce conditional variances, tail probabilities, and other features of within-group distributions. This is especially relevant in economics, where variation around the mean is often itself the object of interest, including in research on inequality, risk, and heterogeneous policy effects (8, 9, 33). More generally, the results raise the question of whether structural fidelity in higher moments is a broader property of LLM-generated data, or whether it depends on the outcome, the conditioning variables, and the amount of information provided to the model.

One major limitation is that structural fidelity becomes less precise at finer levels of detail. Exploratory analyses of distributional tails show that the models broadly recover where very low and very high life satisfaction are concentrated across the income distribution, but often misestimate their prevalence. Country-level correspondence is also only moderate: rank correlations between human and LLM country-specific gradients range from 0.54 to 0.80 across the 65-country analysis and are relatively insensitive to retaining Iraq, whereas Pearson correlations are considerably more sensitive to this influential observation. Structural fidelity thus appears strongest as a broad population-level pattern rather than as an exact representation of extreme outcomes or country-specific gradients.

Several other limitations remain. First, because we obtain only one prediction per profile, the observed dispersion combines between-profile variation with any stochastic variation arising from a single generation. We cannot separate these components or determine how much within-profile variation would be observed under repeated sampling. This is particularly relevant for comparisons of absolute dispersion across models because generation parameters were not harmonized across providers and their effective default settings may differ. Cross-model differences in the degree of compression should therefore be interpreted cautiously. Repeated predictions for the same respondent profiles would help separate systematic between-profile variation from within-profile generation variability. Second, household income is measured by respondents' subjective position in their country's income distribution rather than by objective income. Future work should examine whether the same pattern holds using continuous measures of personal income, which would allow more precise estimation of how well-being dispersion changes across the income distribution. Third, the follow-up analyses show that the preserved dispersion gradient does not depend solely on the explicit income cue, but the remaining profiles still contain several characteristics correlated with socioeconomic position, including education, employment, occupation, material hardship, subjective social class, and perceived control. We therefore cannot identify which specific characteristics, or combinations of characteristics, generate the remaining structure. The full counterfactual income-sweep experiment was conducted only with Claude Sonnet 5; Gemini and DeepSeek were used only for the income-ablation replication. Fourth, we chose life satisfaction because previous research has shown that its dispersion tends to decline with income. It remains unclear whether LLMs also reproduce patterns of dispersion in other outcomes, such as political attitudes, health, consumption, or labor-market behavior. Finally, the analyses of distributional tails, country-level gradients, generalization beyond income, and the Gemini and DeepSeek income-ablation replications were exploratory and should be interpreted with care.

More generally, this paper shows that compression in LLM-generated data does not necessarily erase the social structure of human heterogeneity. Across six models, well-being inequality is substantially compressed, but much of its socioeconomic pattern remains visible across income, employment, education, and perceived control. This does not make synthetic LLM populations reliable substitutes for human data, especially when absolute levels, tail probabilities, or subgroup magnitudes matter. Researchers should therefore continue to carefully validate synthetic LLM data for use in the social sciences (37). At the same time, substantial compression can coexist with meaningful preservation of how human heterogeneity is distributed across social groups. Distinguishing the scale of heterogeneity from its structure may thus offer a more useful basis for evaluating synthetic populations and deciding where LLM-generated data can or cannot inform social science research.

## Methods and materials

### Data and sample

We used individual-level data from Wave 7 of the World Values Survey (WVS)[38], a cross-national survey of social values, attitudes, and demographic characteristics. The analysis used the WVS Wave 7 cross-national dataset and retained all respondents with valid responses to the two variables required for the primary analysis: life satisfaction (Q49) and household income position (Q288). No additional sample inclusion or exclusion criteria were imposed. The resulting analysis sample contains 93,901 respondents from 66 countries and territories.

Life satisfaction was measured using Q49: "All things considered, how satisfied are you with your life as a whole these days?", recorded on a scale from 1 ("completely dissatisfied") to 10 ("completely satisfied"). Household income position was measured using Q288, which asks respondents to place their household on a 10-point income scale ranging from the lowest income group (1) to the highest income group (10) in their country, taking account of all wages, salaries, pensions, and other household income. Q288 therefore captures respondents' position in their country's household-income distribution rather than an objectively measured income decile.

The WVS microdata and questionnaire documentation are available from the World Values Survey. The underlying microdata are not redistributed in our repository.

**Construction of respondent profiles**

Following prior work that used the respondent-profile approach on LLM prediction of life satisfaction (11), we constructed a moderately rich text profile for each respondent from the demographic, socioeconomic, and attitudinal WVS variables used as inputs to the models. Each profile began with the respondent's age, sex, and country, followed by all nonmissing responses among the selected survey variables. Missing responses were omitted rather than imputed. Categorical responses were presented using both the original numeric response and its verbal label. The earlier study similarly used natural-language profiles constructed from individual sociodemographic, attitudinal, and psychological characteristics to generate respondent-level life-satisfaction predictions.

The input variables covered family, friends, leisure, politics, work, and religion (Q1–Q6); perceived freedom and control (Q48); recent material hardship (Q51–Q56); generalized trust (Q57); confidence in the press, police, government, political parties, universities, and elections (Q66, Q69, Q71, Q72, Q75, and Q76); views on income equality (Q106); perceived corruption (Q112); neighborhood security (Q131); importance of God and religious-service attendance (Q164 and Q171); importance of democratic government (Q250); national pride (Q254); sex and age (Q260 and Q262); immigrant and citizenship status (Q263 and Q269); household size (Q270); marital status and number of children (Q273 and Q274); education (Q275); employment status and occupation (Q279 and Q281); employment sector (Q284); chief-wage-earner status (Q285); subjective social class (Q287); household income position (Q288);

religious denomination (Q289); and settlement type. Only nonmissing values entered a respondent's profile.

To prevent outcome leakage, happiness (Q46) and life satisfaction (Q49) were excluded from every profile. The preparation code also checked that no profile text contained Q46, Q49, or direct references to life satisfaction before any API calls were made. Household income position (Q288) was retained because income was one of the observed respondent characteristics on which the life-satisfaction predictions were conditioned. Each respondent was assigned a unique WVS row identifier so that LLM predictions could subsequently be merged back to the original survey record.

**LLM prediction procedure**

We generated one life-satisfaction prediction for each respondent profile using six large language models: GPT-5.6 Luna (`gpt-5.6-luna`), Claude Sonnet 5 (`claude-sonnet-5`), Gemini 3.8 Flash (`gemini-3.8-flash`), DeepSeek V4.1 Flash (`deepseek-flash`), Qwen 3.7 Plus (`qwen3.7-plus`), and Gemma 4 31B (`google/gemma-4-31B-it-turbo`, served through DeepInfra). We collected LLM predictions between 20 and 23 September 2026.

All six models received the same respondent profile, the same system instruction, and the same life-satisfaction question. The system instruction stated: "You will be given a World Values Survey respondent profile. Answer the life-satisfaction question as that person would most likely answer it. Return ONLY one integer from 1 to 10. Do not explain your answer." The user prompt then presented the respondent profile followed by the original WVS life-satisfaction question and its 1–10 response scale.

No temperature or top-p parameter was specified for any model. GPT-5.6 Luna was run with reasoning disabled, while thinking was explicitly disabled for DeepSeek V4.1 Flash and Qwen 3.7 Plus. Gemini 3.8 Flash was left at its provider-default thinking setting. Model-specific API routes, output-token limits, and reasoning or thinking settings are reported in Table S5.

We retained one valid prediction per respondent per model. Responses were parsed as valid only if a unique integer from 1 to 10 could be recovered. Failed or unparsable calls were attempted up to six times with exponential backoff. Predictions were processed in batches with periodic checkpointing, allowing interrupted runs to resume without repeating previously completed respondents. After each full model run, we verified that all 93,901 respondent identifiers were unique, that every call had completed successfully, and that all predictions were integer values between 1 and 10. The exact model identifiers, API configuration, prompting code, retry logic, and validation procedures are provided in the public repository.

### Primary measures of well-being inequality

Following Klein Teeselink and Zauberman (20), we measured well-being inequality using the within-income-group standard deviation of life satisfaction. For each of the ten income groups, we calculated the mean and standard deviation of life satisfaction separately for the human responses and for each LLM.

For source $m$, we summarized overall within-income dispersion, $S_m$, as

$$S_m = \sum_{j=1}^{10} w_j SD_{jm},$$

where $SD_{jm}$ is the standard deviation of life satisfaction in income group $j$, and $w_j$ is the proportion of human respondents in that income group. The same human income-group weights were used for the human data and for every LLM. This quantity provides the scale term used in the normalized analysis described below. Because each LLM generated one prediction for each respondent profile, the observed dispersion combines systematic variation across respondent profiles with any stochastic variation arising from the model and provider-default generation settings. Our design does not separately identify these two components.

For the raw income gradients reported in Fig. 1, the mean life-satisfaction slope was estimated by respondent-level OLS of life satisfaction on household income position, while the dispersion slope was estimated as an equally weighted linear trend across the ten income-group standard deviations. Confidence intervals were obtained from 2,000 country-cluster bootstrap replications, with countries resampled with replacement and all respondents within each sampled country retained together.

### Normalized structural-fidelity analysis

Our primary analysis asked whether LLMs preserve the shape of the human income gradient in well-being inequality after removing differences in overall dispersion. For source $m$ and income group $j$, we defined normalized dispersion, $D_{jm}$, as

$$D_{jm} = \frac{SD_{jm}}{\sum_{k=1}^{10} w_k SD_{km}},$$

where $SD_{jm}$ is the standard deviation of life satisfaction for income group $j$ from source $m$, and $w_k$ is the human sample share in income group $k$. The same human income-group weights were used for the human data and for every LLM. By construction, the human-share-weighted mean of $D_{jm}$ equals 1 for each source. Values above 1 therefore indicate greater dispersion than that source's own weighted-average dispersion, while values below 1 indicate lower dispersion.

This normalization removes each source's overall dispersion scale while preserving the relative pattern of heterogeneity across income groups. Under pure proportional compression, where

$$SD_{j,\text{LLM}} = cSD_{j,\text{Human}}$$

For a constant $c$, the compression factor cancels and the normalized human and LLM profiles are identical.

Inference for the primary profile comparison used a paired country-cluster bootstrap (1,000 replications). Countries were resampled with replacement, and the human and all six LLM outcomes for respondents from each sampled country were carried together within each bootstrap replication. The original human income-group weights were held fixed across replications. For each replication, we recalculated the ten group-specific standard deviations and normalized profiles for all seven sources.

For each LLM, we tested equality of its full ten-category normalized profile with the human profile using a global Wald test based on the bootstrap covariance matrix of the ten model-minus-human differences. Because normalization imposes one linear constraint on the ten-category profile, the covariance matrix has rank nine. As a summary of the overall gradient, we also fitted a linear trend across income groups 1–10 to each normalized profile and reported the two-sided bootstrap confidence interval and $p$-value for the difference between the LLM and human slopes. The full-profile Wald test, rather than the linear-trend comparison, was the primary confirmatory test.

## Income-cue mechanism experiment

To examine whether the preserved income-dispersion gradient was driven directly by the income information supplied to the model or by other information in the respondent profile, we conducted a separately preregistered follow-up experiment using Claude Sonnet 5. Claude provided a particularly informative case because its normalized linear income-dispersion slope in the main analysis was almost identical to the human slope despite substantial compression in absolute dispersion. We drew a stratified random sample of 2,000 respondents, with 200 respondents from each of the ten household-income groups.

We implemented two manipulations. In the counterfactual income sweep, each respondent was assigned each of the ten possible income positions in turn, producing 20,000 profiles, while every other profile characteristic remained unchanged. In the income-ablation condition, household income position (Q288) was removed from the profile while all other information was retained. Predictions from the original observed-income condition for the same 2,000 respondents were used as the comparison condition.

Normalized dispersion was calculated using the income-group shares from the full human sample rather than the equal sampling shares in the experimental subsample. For the counterfactual sweep, we estimated linear trends in mean predicted life satisfaction and normalized dispersion across assigned income positions. The quadratic specification and endpoint-to-minimum dispersion ratio were added post-preregistration as descriptive analyses. For the income-ablation analysis, respondents were grouped by their observed WVS income position, and the observed-income and no-income dispersion slopes were compared using 1,000 paired country-cluster bootstrap replications.

As a post-preregistration generalization check, we repeated the income-ablation analysis for Gemini 3.8 Flash and DeepSeek V4.1 Flash using the same respondents, profile construction, normalization, and bootstrap procedure. The counterfactual income sweep was not repeated for these models.

## Country-fixed-effect RIF regressions

As a robustness check, we examined whether the income gradient in well-being inequality remained after accounting for differences in country composition. We estimated recentered influence function (RIF) regressions for the variance of life satisfaction separately for the human data and for each LLM. Household income position (Q288) entered linearly from 1 to 10, and all models included country fixed effects.

For variance, the RIF is given by

$$RIF(y{:}\,\sigma^2) = (y - \mu)^2,$$

so that regressing the variance RIF on income estimates how the unconditional variance of life satisfaction changes with household income position after accounting for country fixed effects. The same specification was applied separately to the observed human life-satisfaction scores and to each model's predictions.

These regressions were intended as a supporting test of whether the negative income–dispersion gradient observed in the pooled data also appears within countries. Because the coefficients are expressed in raw variance units, their magnitudes remain affected by each LLM's overall degree of variance compression and should not be interpreted as scale-adjusted measures of structural fidelity.

A supplementary categorical specification treated income group as a set of indicators, with income group 1 as the reference category, to examine departures from linearity in the within-country gradient.

**RIF-Oaxaca decomposition**

As a supplementary robustness analysis, we used an Oaxaca-style decomposition (34, 35) based on recentered influence functions (RIFs) for variance (36) to examine whether differences in life-satisfaction variance between adjacent income groups could be attributed to observed differences in respondent composition. The decomposition was estimated separately for the human data and for each LLM across all nine adjacent income-group transitions.

For each comparison, we estimated the variance RIF in the lower-income group as a function of age, sex, marital status, education, number of children, and country fixed effects, and used the fitted model to construct a counterfactual for the adjacent higher-income group. The resulting variance difference was decomposed into a composition component associated with these observed characteristics and a residual component. Comparisons were restricted to countries represented in both income groups. Confidence intervals were obtained from 1,000 country-cluster bootstrap replications.

**Additional exploratory analyses**

We conducted three sets of exploratory analyses to examine whether the main structural-fidelity result extended beyond the preregistered income analysis. First, we examined distributional tails by comparing the prevalence of very low and very high life satisfaction across income groups in the human and LLM-generated responses. Very low life satisfaction was defined as a score of 3 or below, and very high life satisfaction as a score of 8 or above. We also examined the prevalence of the maximum response of 10 as a supplementary upper-tail diagnostic.

Second, we examined cross-country variation in the strength of the income–dispersion gradient. For each country and source, we calculated the normalized income-specific dispersion profile using the same normalization principle as in the pooled analysis and summarized its gradient across income groups. We then compared the resulting country-specific human and LLM gradients using Pearson and Spearman correlations and cross-country linear regressions. Iraq was excluded from the main country-level comparison because Q49 life satisfaction and Q288 income position were nearly identical in the Iraqi data, producing an extreme and highly influential human gradient.

Third, we examined whether structural fidelity generalized beyond income to employment status (Q279), education (Q275), marital status (Q273), and perceived freedom and control (Q48). For employment, the targeted contrast compared unemployed respondents (code 7) with those employed full-time, employed part-time, or self-employed (codes 1–3). For education, we compared respondents with lower-secondary education or less (codes 0–2) with those with tertiary education (codes 5–8). For marital status, we compared divorced, separated, widowed, or single respondents (codes 3–6) with married or cohabiting respondents (codes 1–2). For perceived freedom and control, we compared low scores (1–3) with high scores (8–10).

For each characteristic, we calculated category-specific standard deviations and normalized them by the source-specific weighted-average standard deviation, using human category shares as fixed weights for humans and all six LLMs. We also calculated the ratio of the two unnormalized standard deviations for each targeted contrast. The reported 95% confidence intervals for these ratios are percentile intervals from 2,000 country-cluster bootstrap replications. Each replication resampled countries with replacement and kept the human and six LLM outcomes together for each respondent. We additionally estimated country-fixed-effect RIF-variance regressions. These analyses used the same respondent-level LLM predictions as the primary income analysis and were treated as exploratory.

**Preregistration and analysis status**

The primary income analysis was preregistered on AsPredicted before the full six-model analysis was conducted. The preregistered confirmatory analysis tested whether the human income gradient in well-being inequality was reproduced by the LLMs after accounting for their overall compression of response variability. The normalized income-dispersion comparison therefore served as the primary confirmatory test. Country-fixed-effect RIF regressions and the RIF-Oaxaca decomposition were specified as supporting robustness analyses. The exact covariate set used in the RIF-Oaxaca implementation was not fully prespecified; the final specification included age, sex, marital status, education, number of children, and country fixed effects. The original preregistration is available at https://aspredicted.org/sf62bj.pdf.

The Claude income-cue experiment was preregistered separately before the full N = 2,000 analysis was conducted (https://aspredicted.org/5hg8ba.pdf). The preregistered analyses included the counterfactual income sweep, the income-ablation comparison, and the linear income-dispersion trends reported in the main text. We conducted a small technical pilot beforehand to test implementation. The quadratic specification used to characterize the U-shaped counterfactual dispersion profile and the endpoint-to-minimum dispersion ratio were added after preregistration and are treated as descriptive analyses.

The Gemini and DeepSeek income-ablation replications were conducted after the Claude preregistration and are treated as exploratory generalization checks. Analyses of distributional tails, cross-country variation in income-dispersion gradients, generalization to employment, education, marital status, and perceived freedom and control, and the OLS and Lasso benchmarks were also conducted post-preregistration and are treated as exploratory or benchmarking analyses.

**Code and data availability**

All code used to construct respondent profiles, generate LLM predictions, reproduce the primary analyses, and conduct the exploratory analyses is available in the public GitHub repository: https://github.com/npowdthavee/LLM_wellbeing_inequality. The repository contains the data-preparation, LLM-prediction, primary-analysis, mechanism-analysis, and generalization notebooks, together with documentation describing the order in which they should be run. The World Values Survey data can be accessed from https://www.worldvaluessurvey.org.

**Artificial intelligence use**

The large language models evaluated in this study were used as research instruments as described above. During manuscript preparation, the author also used OpenAI ChatGPT (GPT-5.6 Sol) to assist with drafting, editing, and improving text clarity. All AI-assisted content was reviewed, verified, and revised by the author, who takes full responsibility for the accuracy and integrity of the manuscript.

## Supplementary Information (SI)

**Table S1.** Primary comparisons of normalized income-dispersion profiles between LLMs and humans

| Model | Global Wald $\chi^2$ | Global p-value | Slope difference (95% CI) | Profile RMSD |
|---|---|---|---|---|
| GPT-5.6 Luna | 335.69 | < .001 | −0.038 [−0.047, −0.030] | 0.147 |
| Claude Sonnet 5 | 119.24 | < .001 | +0.003 [−0.006, 0.013] | 0.100 |
| Gemini 3.8 Flash | 136.25 | < .001 | −0.014 [−0.021, −0.006] | 0.075 |
| DeepSeek V4.1 Flash | 68.06 | < .001 | +0.001 [−0.007, 0.009] | 0.051 |
| Gemma 4 31B | 87.76 | < .001 | −0.026 [−0.033, −0.018] | 0.079 |
| Qwen 3.7 Plus | 164.92 | < .001 | −0.030 [−0.038, −0.023] | 0.107 |

**Note**. Global Wald tests compare each LLM's full normalized ten-group income-dispersion profile with the human profile, using the covariance matrix from 1,000 paired country-cluster bootstrap replications. All tests have 9 degrees of freedom. Slope differences are the LLM minus human linear trend; negative values indicate a steeper decline than in the human data. Brackets contain 95% percentile bootstrap confidence intervals. Profile RMSD is the root-mean-square deviation from the human normalized profile; smaller values indicate closer agreement.

**Table S2.** Robustness of normalized income-dispersion slope differences to alternative weighting schemes

| Model | Primary | Equal-country | WVS survey-weighted |
|---|---|---|---|
| GPT-5.6 Luna | -0.038 | -0.040 | -0.037 |
| Claude Sonnet 5 | +0.003 | +0.001 | +0.005 |
| Gemini 3.8 Flash | -0.014 | -0.015 | -0.013 |
| DeepSeek V4.1 Flash | +0.001 | -0.002 | +0.001 |
| Gemma 4 31B | -0.026 | -0.026 | -0.025 |
| Qwen 3.7 Plus | -0.030 | -0.032 | -0.030 |

**Note.** Entries are differences between each LLM's normalized linear income-dispersion slope and the corresponding human slope. The primary specification uses the original respondent-level sample. Equal-country weighting gives each country equal total weight. The WVS survey-weighted specification uses W_WEIGHT, available for all 93,901 observations. Negative values indicate a steeper normalized decline than in the human data.

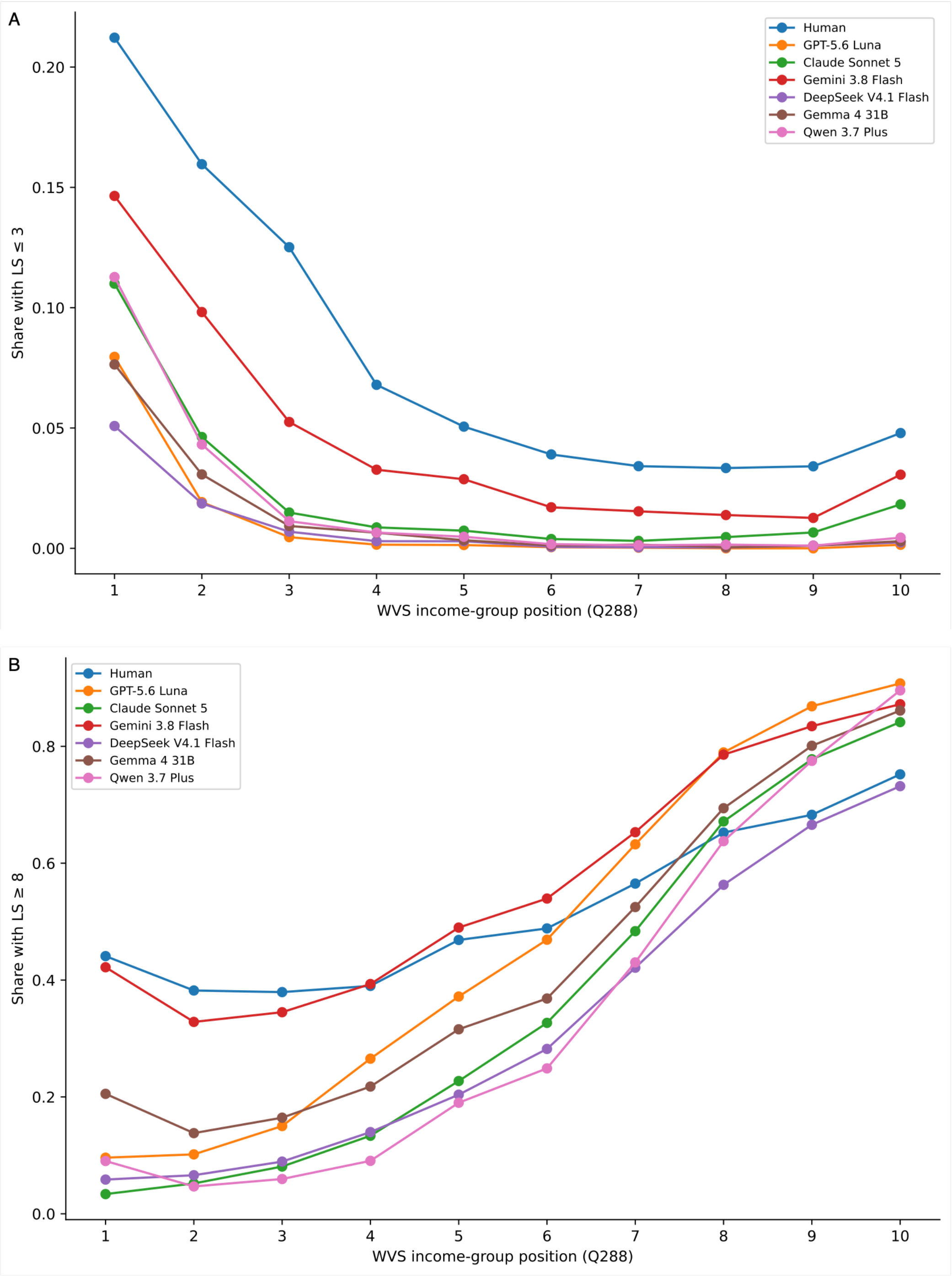

**Fig S1. Income gradients in the lower and upper tails of the life-satisfaction distribution.** The figure shows the prevalence of very low and very high life satisfaction across household income groups for human respondents and the six LLMs. Very low life satisfaction is defined as a score of 3 or below on the 1–10 WVS life-satisfaction scale, and very high life satisfaction as a score of 8 or above. Estimates are calculated separately within each household income group using the same respondent sample as the main analyses. Lines connect income-group estimates for visual presentation.

**Table S3.** RIF-Oaxaca decomposition of adjacent-income changes in life-satisfaction variance with 95% confidence intervals

| Source | Transition | Total Δ (95% CI) | Composition (95% CI) | Residual (95% CI) |
|---|---|---|---|---|
| Human | 1→2 | -3.283 [-3.990, -2.580] | -0.703 [-1.328, -0.012] | -2.580 [-3.148, -2.044] |
| Human | 2→3 | -1.108 [-1.468, -0.746] | -0.068 [-0.358, +0.316] | -1.040 [-1.452, -0.673] |
| Human | 3→4 | -0.904 [-1.181, -0.663] | -0.149 [-0.254, -0.050] | -0.755 [-1.014, -0.534] |
| Human | 4→5 | +0.018 [-0.230, +0.260] | +0.159 [+0.003, +0.331] | -0.141 [-0.332, +0.022] |
| Human | 5→6 | -0.980 [-1.230, -0.713] | -0.419 [-0.622, -0.216] | -0.561 [-0.724, -0.411] |
| Human | 6→7 | -0.255 [-0.432, -0.051] | -0.126 [-0.242, -0.010] | -0.130 [-0.300, +0.065] |
| Human | 7→8 | +0.090 [-0.172, +0.355] | +0.083 [-0.062, +0.221] | +0.007 [-0.235, +0.249] |
| Human | 8→9 | +0.173 [-0.328, +0.663] | -0.098 [-0.336, +0.112] | +0.272 [-0.146, +0.673] |
| Human | 9→10 | +1.082 [+0.240, +2.298] | +0.139 [-0.391, +0.732] | +0.943 [+0.293, +1.812] |
| GPT-5.6 Luna | 1→2 | -0.874 [-1.011, -0.728] | +0.017 [-0.148, +0.195] | -0.891 [-1.049, -0.738] |
| GPT-5.6 Luna | 2→3 | -0.358 [-0.429, -0.278] | -0.040 [-0.088, +0.017] | -0.318 [-0.391, -0.248] |
| GPT-5.6 Luna | 3→4 | -0.148 [-0.204, -0.096] | -0.034 [-0.060, -0.009] | -0.114 [-0.159, -0.072] |
| GPT-5.6 Luna | 4→5 | -0.031 [-0.074, +0.015] | -0.019 [-0.038, +0.006] | -0.012 [-0.049, +0.027] |
| GPT-5.6 Luna | 5→6 | -0.199 [-0.255, -0.151] | -0.072 [-0.117, -0.037] | -0.127 [-0.164, -0.096] |
| GPT-5.6 Luna | 6→7 | -0.151 [-0.184, -0.121] | -0.008 [-0.030, +0.013] | -0.143 [-0.174, -0.113] |
| GPT-5.6 Luna | 7→8 | -0.111 [-0.139, -0.081] | +0.001 [-0.015, +0.017] | -0.112 [-0.141, -0.082] |
| GPT-5.6 Luna | 8→9 | +0.017 [-0.044, +0.078] | +0.001 [-0.025, +0.026] | +0.016 [-0.042, +0.075] |
| GPT-5.6 Luna | 9→10 | +0.527 [+0.344, +0.748] | -0.012 [-0.043, +0.031] | +0.539 [+0.366, +0.756] |
| Claude Sonnet 5 | 1→2 | -0.625 [-0.772, -0.466] | +0.076 [-0.086, +0.278] | -0.701 [-0.903, -0.518] |
| Claude Sonnet 5 | 2→3 | -0.434 [-0.543, -0.321] | -0.055 [-0.104, -0.003] | -0.379 [-0.482, -0.277] |
| Claude Sonnet 5 | 3→4 | -0.190 [-0.241, -0.138] | -0.017 [-0.041, +0.005] | -0.173 [-0.217, -0.124] |
| Claude Sonnet 5 | 4→5 | -0.076 [-0.127, -0.017] | -0.008 [-0.029, +0.012] | -0.068 [-0.118, -0.015] |
| Claude Sonnet 5 | 5→6 | -0.116 [-0.165, -0.069] | -0.044 [-0.076, -0.016] | -0.071 [-0.113, -0.034] |
| Claude Sonnet 5 | 6→7 | -0.010 [-0.061, +0.035] | +0.018 [+0.002, +0.034] | -0.029 [-0.077, +0.013] |
| Claude Sonnet 5 | 7→8 | +0.045 [-0.020, +0.117] | +0.009 [-0.009, +0.030] | +0.036 [-0.026, +0.105] |
| Claude Sonnet 5 | 8→9 | +0.143 [+0.015, +0.284] | -0.011 [-0.053, +0.032] | +0.154 [+0.033, +0.290] |
| Claude Sonnet 5 | 9→10 | +0.492 [+0.215, +0.835] | +0.103 [-0.042, +0.299] | +0.389 [+0.108, +0.700] |
| Gemini 3.8 Flash | 1→2 | -2.905 [-3.441, -2.305] | -0.079 [-0.707, +0.787] | -2.826 [-3.566, -2.199] |
| Gemini 3.8 Flash | 2→3 | -1.014 [-1.299, -0.727] | -0.132 [-0.268, +0.027] | -0.882 [-1.146, -0.631] |
| Gemini 3.8 Flash | 3→4 | -0.426 [-0.594, -0.277] | -0.088 [-0.177, -0.013] | -0.338 [-0.480, -0.196] |
| Gemini 3.8 Flash | 4→5 | +0.119 [-0.057, +0.326] | +0.046 [-0.054, +0.154] | +0.073 [-0.070, +0.225] |
| Gemini 3.8 Flash | 5→6 | -0.842 [-1.035, -0.632] | -0.301 [-0.442, -0.165] | -0.542 [-0.665, -0.429] |
| Gemini 3.8 Flash | 6→7 | -0.244 [-0.398, -0.084] | -0.022 [-0.098, +0.060] | -0.222 [-0.391, -0.072] |
| Gemini 3.8 Flash | 7→8 | -0.051 [-0.211, +0.111] | +0.058 [-0.023, +0.130] | -0.108 [-0.259, +0.058] |
| Gemini 3.8 Flash | 8→9 | +0.122 [-0.198, +0.502] | -0.061 [-0.218, +0.078] | +0.183 [-0.117, +0.525] |
| Gemini 3.8 Flash | 9→10 | +1.379 [+0.625, +2.144] | +0.418 [+0.048, +0.846] | +0.961 [+0.228, +1.706] |
| DeepSeek V4.1 Flash | 1→2 | -0.335 [-0.433, -0.236] | +0.048 [-0.045, +0.158] | -0.383 [-0.513, -0.272] |
| DeepSeek V4.1 Flash | 2→3 | -0.215 [-0.291, -0.143] | -0.035 [-0.090, +0.027] | -0.180 [-0.250, -0.120] |
| DeepSeek V4.1 Flash | 3→4 | -0.100 [-0.159, -0.047] | -0.012 [-0.039, +0.011] | -0.088 [-0.140, -0.039] |

| Source | Transition | Total Δ (95% CI) | Composition (95% CI) | Residual (95% CI) |
|---|---|---|---|---|
| DeepSeek V4.1 Flash | 4→5 | -0.011 [-0.066, +0.038] | -0.009 [-0.039, +0.023] | -0.003 [-0.046, +0.038] |
| DeepSeek V4.1 Flash | 5→6 | -0.253 [-0.312, -0.199] | -0.066 [-0.108, -0.025] | -0.187 [-0.219, -0.157] |
| DeepSeek V4.1 Flash | 6→7 | -0.094 [-0.141, -0.049] | +0.001 [-0.024, +0.026] | -0.095 [-0.136, -0.056] |
| DeepSeek V4.1 Flash | 7→8 | -0.032 [-0.083, +0.020] | +0.015 [-0.005, +0.035] | -0.047 [-0.092, -0.0004] |
| DeepSeek V4.1 Flash | 8→9 | +0.035 [-0.060, +0.149] | +0.001 [-0.046, +0.044] | +0.034 [-0.051, +0.134] |
| DeepSeek V4.1 Flash | 9→10 | +0.261 [+0.089, +0.467] | +0.049 [-0.057, +0.192] | +0.212 [+0.030, +0.425] |
| Gemma 4 31B | 1→2 | -0.838 [-0.970, -0.662] | +0.090 [-0.097, +0.359] | -0.928 [-1.146, -0.727] |
| Gemma 4 31B | 2→3 | -0.394 [-0.469, -0.317] | -0.017 [-0.077, +0.053] | -0.377 [-0.450, -0.301] |
| Gemma 4 31B | 3→4 | -0.215 [-0.274, -0.153] | -0.021 [-0.044, +0.004] | -0.194 [-0.242, -0.144] |
| Gemma 4 31B | 4→5 | -0.086 [-0.146, -0.024] | 0.000 [-0.029, +0.032] | -0.086 [-0.143, -0.032] |
| Gemma 4 31B | 5→6 | -0.241 [-0.300, -0.186] | -0.065 [-0.118, -0.019] | -0.177 [-0.209, -0.145] |
| Gemma 4 31B | 6→7 | -0.107 [-0.158, -0.050] | +0.004 [-0.018, +0.028] | -0.111 [-0.159, -0.062] |
| Gemma 4 31B | 7→8 | -0.076 [-0.130, -0.020] | +0.017 [-0.004, +0.042] | -0.094 [-0.145, -0.038] |
| Gemma 4 31B | 8→9 | +0.004 [-0.094, +0.112] | +0.022 [-0.045, +0.110] | -0.018 [-0.140, +0.108] |
| Gemma 4 31B | 9→10 | +0.308 [+0.148, +0.503] | +0.056 [-0.045, +0.193] | +0.252 [+0.050, +0.476] |
| Qwen 3.7 Plus | 1→2 | -0.962 [-1.084, -0.813] | +0.021 [-0.130, +0.239] | -0.983 [-1.168, -0.794] |
| Qwen 3.7 Plus | 2→3 | -0.516 [-0.599, -0.431] | -0.004 [-0.054, +0.063] | -0.512 [-0.589, -0.426] |
| Qwen 3.7 Plus | 3→4 | -0.232 [-0.288, -0.178] | -0.030 [-0.053, -0.007] | -0.203 [-0.246, -0.158] |
| Qwen 3.7 Plus | 4→5 | -0.050 [-0.106, +0.003] | -0.004 [-0.029, +0.025] | -0.045 [-0.094, -0.001] |
| Qwen 3.7 Plus | 5→6 | -0.230 [-0.283, -0.179] | -0.061 [-0.104, -0.024] | -0.169 [-0.207, -0.133] |
| Qwen 3.7 Plus | 6→7 | -0.044 [-0.096, +0.001] | +0.010 [-0.008, +0.029] | -0.054 [-0.102, -0.010] |
| Qwen 3.7 Plus | 7→8 | -0.017 [-0.067, +0.034] | +0.007 [-0.009, +0.024] | -0.025 [-0.075, +0.026] |
| Qwen 3.7 Plus | 8→9 | +0.032 [-0.053, +0.122] | -0.007 [-0.046, +0.032] | +0.038 [-0.041, +0.126] |
| Qwen 3.7 Plus | 9→10 | +0.295 [+0.145, +0.481] | +0.005 [-0.063, +0.101] | +0.290 [+0.116, +0.474] |

**Note.** Entries report point estimates with 95% percentile confidence intervals from 1,000 country-cluster bootstrap replications. The composition component captures the portion associated with differences in age, sex, marital status, education, number of children, and country composition; the residual component is the portion not attributable to these observed characteristics. Total change equals composition plus residual. Each adjacent comparison is restricted to countries represented in both income groups. Negative values indicate a decline in life-satisfaction variance from the lower to the higher income group; positive values indicate an increase.

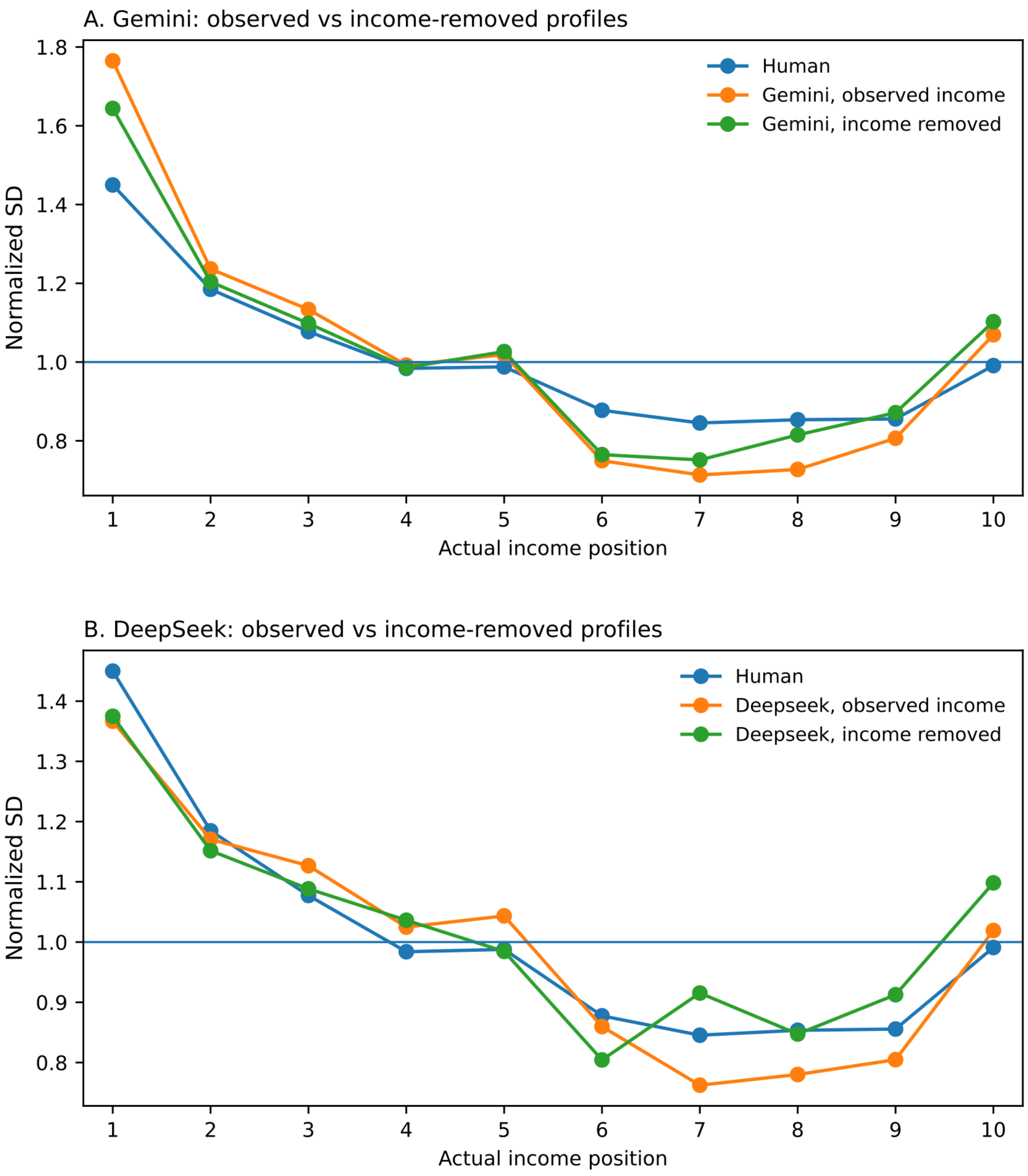

**Fig. S2. Income-ablation results for Gemini 3.8 Flash and DeepSeek V4.1 Flash.** The figure compares the human dispersion profile with each model's predictions when income is included versus removed. Normalized standard deviations use the same human income-group weights as in the main analysis.

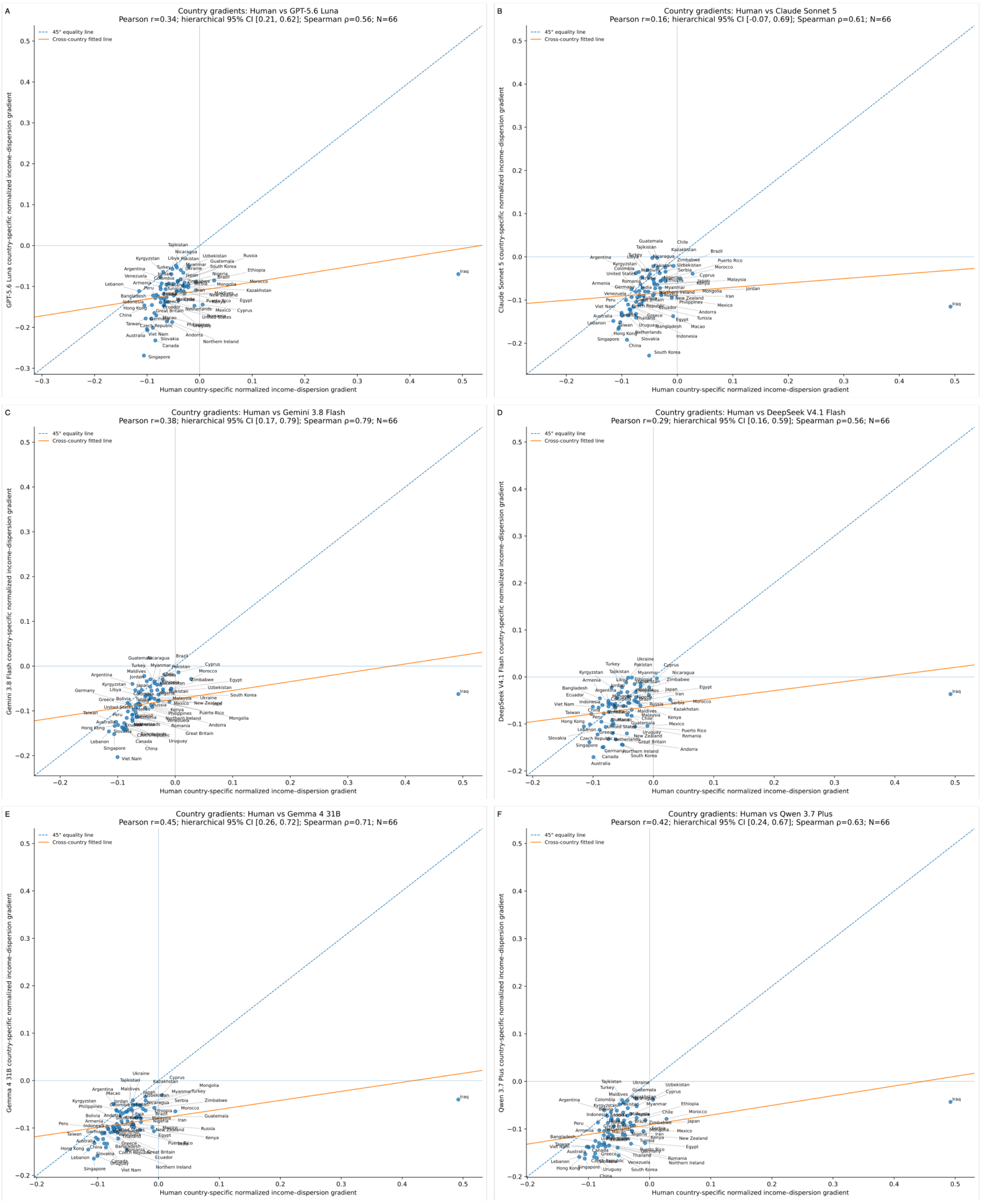

**Fig S3**. **Country-level fidelity in income gradients in well-being inequality, including Iraq.** Each panel compares the country-specific human income gradient in normalized life-satisfaction dispersion with the corresponding gradient produced by one LLM for all 66 countries and territories. Each point represents one country or territory. The dashed 45° line indicates equality between human and LLM gradients, and the solid fitted line shows the cross-country linear relationship. Iraq is retained in these analyses and appears as an extreme outlier in the human country-specific gradient. Pearson and Spearman correlations shown in each panel are calculated across all 66 countries and territories.

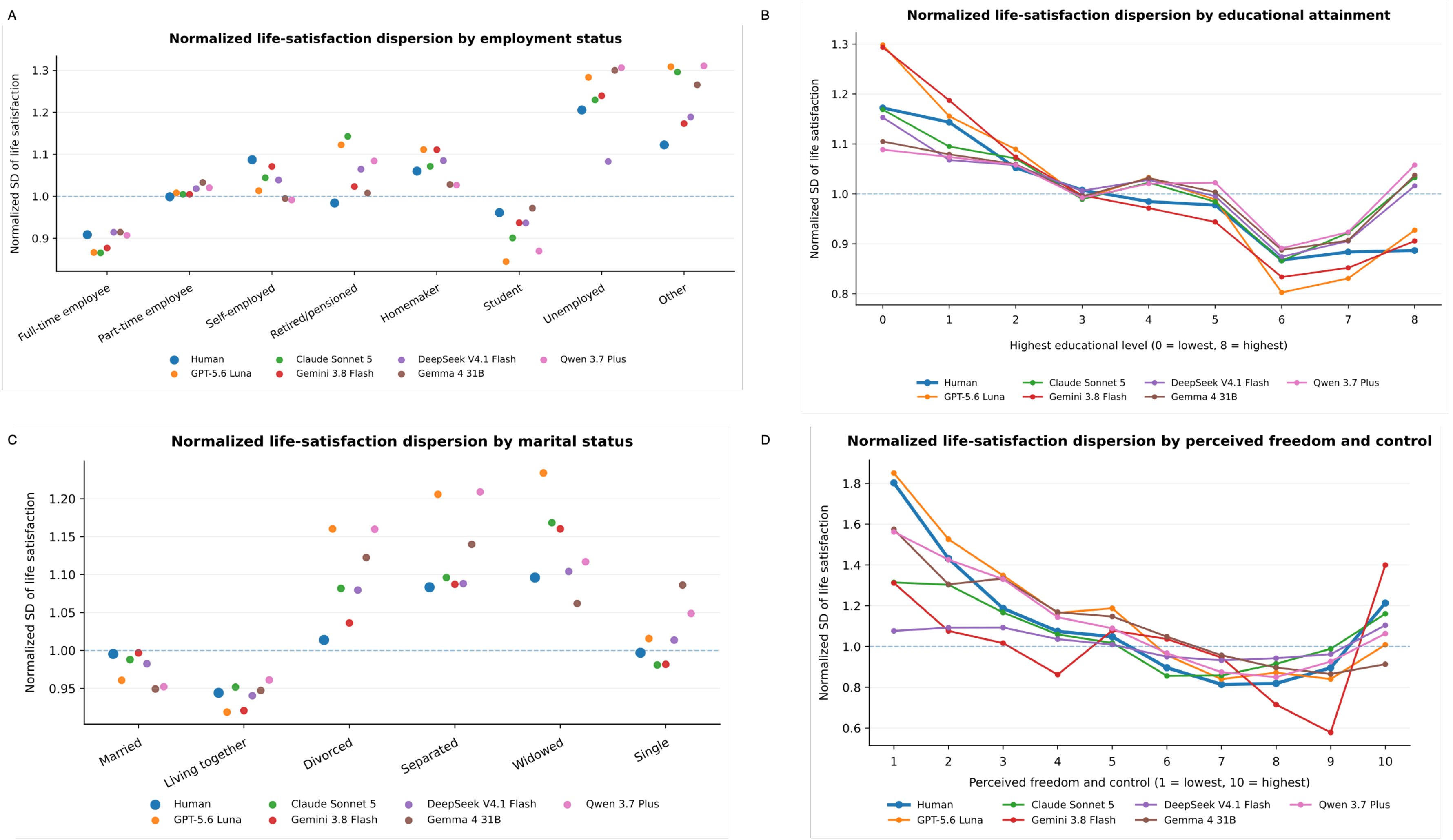

**Fig S4. Normalized well-being inequality across other socioeconomic and psychosocial characteristics.** The four panels show normalized within-group standard deviations of life satisfaction by employment status, educational attainment, marital status, and perceived freedom and control. For each characteristic and source, within-group standard deviations are divided by the source-specific weighted-average standard deviation across categories, using the corresponding human category shares as fixed weights for humans

and all six LLMs. A value of 1 represents the source-specific weighted average for that characteristic. Values above or below 1 indicate relatively greater or lower well-being inequality within that category. The figure extends the scale-adjusted comparison used in Fig. 2 beyond household income.

**Table S4.** Structural fidelity of LLM and supervised statistical predictions relative to the human income-dispersion profile

| Source | Normalized slope | Difference from human | Profile RMSD | Source type |
|---|---|---|---|---|
| **Human** | **-0.049** | **+0.000** | **0.000** | **Human benchmark** |
| GPT-5.6 Luna | -0.087 | -0.038 | 0.147 | LLM |
| Claude Sonnet 5 | -0.046 | +0.003 | 0.100 | LLM |
| Gemini 3.8 Flash | -0.063 | -0.014 | 0.075 | LLM |
| DeepSeek V4.1 Flash | -0.048 | +0.001 | 0.051 | LLM |
| Gemma 4 31B | -0.075 | -0.026 | 0.079 | LLM |
| Qwen 3.7 Plus | -0.079 | -0.030 | 0.107 | LLM |
| OLS | -0.037 | +0.012 | 0.081 | Supervised benchmark |
| Lasso | -0.037 | +0.012 | 0.079 | Supervised benchmark |

**Note.** Normalized slopes are estimated across the ten income groups. The difference-from-human column reports the model slope minus the human slope. Profile RMSD is the root-mean-square deviation between each model's normalized ten-group income-dispersion profile and the human profile; lower values indicate closer structural fidelity. OLS and Lasso predictions are five-fold out-of-sample predictions from supervised models estimated using observed WVS life satisfaction as the training outcome. The LLMs were not given life-satisfaction outcomes from the study sample. OLS/Lasso analyses were conducted post-preregistration as a benchmark.

**Table S5.** Model-specific API request settings for life-satisfaction predictions

| Model | API route | Model identifier | Output cap | Reasoning or thinking |
|---|---|---|---|---|
| GPT-5.6 Luna | OpenAI Responses | gpt-5.6-luna | 16 tokens | Reasoning effort: none |
| Claude Sonnet 5 | Anthropic Messages | claude-sonnet-5 | 16 tokens | No setting specified |
| Gemini 3.8 Flash | Google Gen AI | gemini-3.8-flash | Not specified | Thinking not specified |
| DeepSeek V4.1 Flash | DeepSeek chat completions | deepseek-flash | 16 tokens | Thinking disabled |
| Qwen 3.7 Plus | Alibaba chat completions | qwen3.7-plus | Not specified | enable_thinking: false |
| Gemma 4 31B | DeepInfra chat completions | google/gemma-4-31B-it-turbo | 16 tokens | No setting specified |

**Note**. Temperature, top-p, and random seed were omitted from all six requests; their effective values need not have been identical across providers. “Not specified” indicates that the parameter was omitted from the request.

The common instruction and respondent prompt were passed through each provider’s corresponding fields: OpenAI instructions/input, Anthropic and Gemini system instruction plus user content, or separate system and user messages for chat completions.